\documentclass[10pt,twocolumn,letterpaper]{article}

\usepackage{cvpr}              
\usepackage{times}
\usepackage{latexsym}
\usepackage{graphicx}
\usepackage{array,booktabs}
\usepackage{colortbl}
\usepackage{amssymb}
\usepackage{amsfonts}
\usepackage{amsmath}
\usepackage{csquotes}
\usepackage{bbm}
\usepackage{pifont}
\newcommand{\xmark}{\text{\ding{55}}}

\usepackage[T1]{fontenc}

\usepackage[utf8]{inputenc}
\usepackage[accsupp]{axessibility}

\usepackage{microtype}

\usepackage{inconsolata}

\usepackage[dvipsnames]{xcolor}

\definecolor{cvprblue}{rgb}{0.21,0.49,0.74}
\usepackage[pagebackref,breaklinks,colorlinks,citecolor=cvprblue]{hyperref}

\def\paperID{21} 
\def\confName{CVPR}
\def\confYear{2024}

\title{PromptSync: Bridging Domain Gaps in Vision-Language Models through Class-Aware Prototype Alignment and Discrimination}

\author{Anant Khandelwal\\
Glance AI\\
{\tt\small anant.iitd.2085@gmail.com}
}

\begin{document}
\maketitle
\begin{abstract}
The potential for zero-shot generalization in vision-language (V-L) models such as CLIP has spurred their widespread adoption in addressing numerous downstream tasks. Previous methods have employed test-time prompt tuning to adapt the model to unseen domains, but they overlooked the issue of imbalanced class distributions. In this study, we explicitly address this problem by employing class-aware prototype alignment weighted by mean class probabilities obtained for the test sample and filtered augmented views. Additionally, we ensure that the class probabilities are as accurate as possible by performing prototype discrimination using contrastive learning. The combination of alignment and discriminative loss serves as a geometric regularizer, preventing the prompt representation from collapsing onto a single class and effectively bridging the distribution gap between the source and test domains. Our method, named PromptSync, synchronizes the prompts for each test sample on both the text and vision branches of the V-L model. In empirical evaluations on the domain generalization benchmark, our method outperforms previous best methods by 2.33\% in overall performance, by 1\% in base-to-novel generalization, and by 2.84\% in cross-dataset transfer tasks.
\end{abstract}    
\section{Introduction}
Training Vision-Language Models (VLMs) with large-scale image-text pairs is known for imparting robust generalization capabilities across diverse downstream tasks \cite{radford2021learning, jia2021scaling, zhai2022lit, yao2021filip, yuan2021florence, alayrac2022flamingo}. However, training these models from scratch for each downstream task is very time-consuming. Moreover, the essence of pre-training with a large-scale dataset is lost when the pre-trained model is not generalizable across downstream tasks. This is due to unexpected changes in data distribution, and the sensitivity to these shifts leads to a decline in performance \cite{hendrycks2019benchmarking, recht2019imagenet, quinonero2008dataset}. To tackle this, there exist three most commonly used techniques: fine-tuning \cite{oquab2014learning}, prompt tuning \cite{zhou2022learning}, adapter \cite{he2021towards}, and LoRA \cite{hu2021lora}. Among these, prompt tuning is the simple, recent, and most widely used technique for foundation models \cite{zhou2022learning, zhou2022conditional, khattak2023maple, khattak2023self, zang2022unified}. However, prompt learning/tuning approaches are used during the training phase to learn representative prompts based on the training data for the downstream task. This approach does not specifically address the distribution shift present in the dataset. Recent methods, TPT \cite{shu2022test} and PromptAlign \cite{samadh2023align}, adjusts the learnable prompt tokens dynamically during testing to enable test-time adaptation and align the context of the test sample as per the seen distribution by the model. Specifically, TPT \cite{shu2022test} updates the learnable prompt tokens (keeping the model parameters frozen) by minimizing the entropy of top-$N_k$ confidently predicted samples, acquired through diverse augmented views of the incoming test sample. Additionally, PromptAlign \cite{samadh2023align} aligns token distribution of the test sample in the visual branch with the pre-computed statistics of the complete proxy source dataset irrespective of the fact that one class distribution may have different mean and variance than the other classes.

In this work, we demonstrate the multi-modal test-time adaptation of prompts. In contrast to PromptAlign, which aligns the distribution for the complete source dataset with test sample, we propose class-aware prototype alignment to address the distributional shift on a class-wise basis. For instance, in an open world there are 360 different breeds of dogs compared to only 71 for cats, leading to one class having higher variance than the others. For each test sample, we obtain randomly augmented views (for both text and image) that are fed to the model for prompt tuning on both the textual and visual branches. We adapt the learnable prompt tokens by aligning the prototype for test sample and confident augmented views with the pre-computed class prototypes (obtained from the proxy source dataset) weighted by the mean probability of each class obtained from confident augmented views. Before alignment, we update the prompt tokens on both the text and visual branches using prototype discrimination and then use updated prompts to align the test sample and augmented views with class prototypes using mean class probabilities. This is based on the idea that prototype vector can capture the complete information of mean and variance for each class distribution and hence it mitigates the class collapse (during test time adaptation) due to high variance of particular classes. Empirical evaluation of our methods shows state-of-the-art Top-1 accuracy for three tasks: domain generalization, base-to-novel generalization, and cross-dataset transfer. This validates the effectiveness of our method in enhancing zero-shot generalization. Our contributions can be summarized as follows:
\begin{itemize}
    \item We propose a class-aware prototype alignment technique for individual test samples to align the context of each test sample with the source distribution on a class-wise basis, thereby mitigating the effects of distributional shift between classes. 
    \item We propose class-aware prototype discrimination to discover the class distribution for efficient alignment. Additionally, we propose the offline computation of class prototypes from a proxy source dataset for foundation V-L models.
    \item We propose multi-modal test-time prompt tuning for both text and visual branches. Empirical evaluation on base-to-novel generalization, domain generalization, and cross-dataset transfer shows the efficiency of our method over existing methods.
\end{itemize}
\section{Related Work}
Vision-Language (V-L) foundation models like CLIP \cite{radford2021learning} and ALIGN \cite{jia2021scaling} have emerged as robust zero-shot generalizable models. They integrate image and text modalities through pre-training on extensive image-text pairs. However, adapting these models to specific downstream tasks with limited data remains challenging. Recent methods explore prompt tuning in CLIP-like models, treating prompts as continuous learnable vectors and fine-tuning them while keeping the model parameters frozen. CoOp \cite{zhou2022learning} proposed fine-tuning CLIP by learning a set of prompts in the text encoder. CoCoOp \cite{zhou2022conditional}, an improvement over CoOp, dynamically conditions the text prompts by the image embeddings. MaPLe \cite{khattak2023maple} is a deep prompting baseline that tunes prompts on both text and image branches, further conditioning image prompts on text prompts using a V-L coupling function. However, these approaches necessitate training data for prompt learning, limiting adaptation to novel datasets during test time. Recent approaches like TPT \cite{shu2022test} aim to learn prompts exclusively at test time but encounter challenges in handling distribution misalignment between CLIP's pre-training data and downstream test data. PromptAlign \cite{samadh2023align} addresses this by introducing token distribution alignment in the image branch. However, it does not account for the potential variance in class distributions. In contrast, our method, inspired by a multi-modal prompting variant \cite{khattak2023maple}, actively aligns class prototypes by leveraging a proxy dataset as a substitute for unavailable CLIP pre-training data. To our knowledge, our approach is the first to explicitly address class-aware distribution misalignment in V-L foundational models during test time.
\section{Methodology}
\textbf{Revisiting CLIP}: Our approach is based on the pre-trained V-L model: Contrastive Language-Image Pre-Training (CLIP). It consists of a text and visual encoder (denoted by $\mathcal{F}_t$ and $\mathcal{F}_v$, respectively, and their pre-trained parameters are represented by $\theta_{\textsc{\tiny{CLIP}}} = \{\theta_t, \theta_v\}$, respectively), used for mapping the text and image to the vector representation, respectively. The input image is $\mathbf{X}$, which is divided into $M$ patches, and the [CLS] token is prepended to these $M$ patch tokens that are projected to produce $\tilde{\mathbf{X}}_{v} = \{\mathbf{e}_{\tiny{\textsc{[CLS]}}}, \mathbf{e}_{1}, \mathbf{e}_{2},...... \mathbf{e}_{M}\}$, where $e_i$ is the embedding for the corresponding patch token in $\mathbf{X}$. The image encoder produces latent visual feature representation $\tilde{\boldsymbol{f}}_{v} = \mathcal{F}_{v}(\tilde{\mathbf{X}}_{v}, \theta_v)$ with transformer blocks from $\tilde{\mathbf{X}}_{v}$. The class label $y$ is embedded within a text template, such as \enquote{\textit{a photo of a \small{<CLS>}}} resulting in $\tilde{\mathbf{X}}_{t} = \{\tiny{SOS}, \mathbf{t}_1, \mathbf{t}_2,..., \mathbf{t}_L, \mathbf{c}_k, \tiny{EOS}\}$, where SOS and EOS are the start and end token embeddings and $\mathbf{t}_l|_{l=1}^{L}$ and $\mathbf{c}_k$ are the token embeddings corresponding to the text template and the class label, respectively. Similarly, the text encoder $\mathcal{F}_t$ encodes $\tilde{\mathbf{X}}_{t}$ with transformer blocks to produce latent text feature representation $\tilde{\boldsymbol{f}}_{t} = \mathcal{F}_{t}(\tilde{\mathbf{X}}_{t}, \theta_t)$. For zero-shot inference, each text feature for class labels $y = \{1, 2,..... C\}$ is paired with an image feature to compute the similarity score $s_i = \textrm{sim}(\tilde{\boldsymbol{f}}_{t_i} \cdot \tilde{\boldsymbol{f}}_{v})$ where $\textrm{sim}(\cdot)$ denotes cosine similarity. The predicted probability on $\textbf{X}$ for each $y_i$ is given as $p(y_i| \textbf{X}) = \frac{e^{\textrm{sim}(\tilde{\boldsymbol{f}}_{t_i} \cdot \tilde{\boldsymbol{f}}_{v})/\tau}}{\sum_{j=1}^{C} e^{ \textrm{sim}(\tilde{\boldsymbol{f}}_{t_j} \cdot \tilde{\boldsymbol{f}}_{v})/\tau}}$, where $\tau$ is the temperature of softmax.\\
\textbf{Prompt Tuning}: CLIP integrates a considerable pool of knowledge derived from its training on millions of image-text pairs characterized by varying degrees of noise. Prompt tuning methods aim to extract the rich features learned by the CLIP model. Recent approaches \cite{zhou2022conditional, zhou2022learning, khattak2023maple, bahng2022visual, zang2022unified} append extra learnable prompts to the input of image and text encoders while keeping them frozen. Modified input prompts with frozen encoders generate undistorted and rich CLIP features, where prompt tuning tries to map the context to the source distribution, i.e., the CLIP pre-training dataset. In our work, we use a recent multi-modal prompting baseline \cite{khattak2023maple} where prompt tuning is performed on both the text and image encoders.
Specifically, the image and text encoders process the input $\tilde{\mathbf{X}}_{v} = \{\mathbf{e}_{\tiny{\textsc{[CLS]}}}, \mathbf{p}_v, \mathbf{e}_{1}, \mathbf{e}_{2}, ...... \mathbf{e}_{M}\}$ and $\tilde{\mathbf{X}}_{t} = \{\tiny{SOS}, \mathbf{p}_t, \mathbf{t}_1, \mathbf{t}_2, ..., \mathbf{t}_L, \mathbf{c}_k, \tiny{EOS}\}$ respectively. The learnable prompts $\textbf{p}_v$ and $\textbf{p}_t$ represent the $V$ visual and $T$ textual tokens, respectively. We will call prompts $\textbf{p}_t$ and $\textbf{p}_v$ as $\textbf{p}$ only. Our approach is based on deep prompting, as in\cite{khattak2023maple}, along with text and image prompts at subsequent transformer blocks. We suggest referring to \cite{khattak2023maple} for more details on baseline architecture.\\
\textbf{Test Time Adaptation}: Test-time adaptation aims to boost generalisation in a zero-shot manner. Existing methods, Test time prompt tuning (TPT)\cite{shu2022test} and PromptAlign\cite{samadh2023align}, both are introduced to provide the model context that is customized for each individual test sample in order to extract rich knowledge from CLIP. For both methods, several augmented views $\mathcal{H}(\textbf{X}_{test})$ are generated from the given test sample $\textbf{X}_{test}$. The average entropy for the filtered views (selected using a confidence threshold) is then used to update the prompts $\textbf{p}$ using the following unsupervised objective:
\begin{equation}
    \mathcal{L}_{ent} = \textrm{arg}\underset{\textbf{p}}{\textrm{min}} - \sum_{i=1}^{C} \tilde{p}_{\textbf{p}}(y_i| \textbf{X}_{test}) \log \tilde{p}_{\textbf{p}}(y_i| \textbf{X}_{test})
    \label{tpt}
\end{equation}
where $\tilde{p}_{\textbf{p}}(y_i| \textbf{X}_{test})$ is the average of vector class probabilities (over the filtered augmented views) produced by the model. Additionally, PromptAlign uses distribution alignment loss, which aligns the mean and variance of filtered augmented views of the test sample with source statistics across layers of the model.
\begin{figure*}
    \centering
    \includegraphics[width=\textwidth]{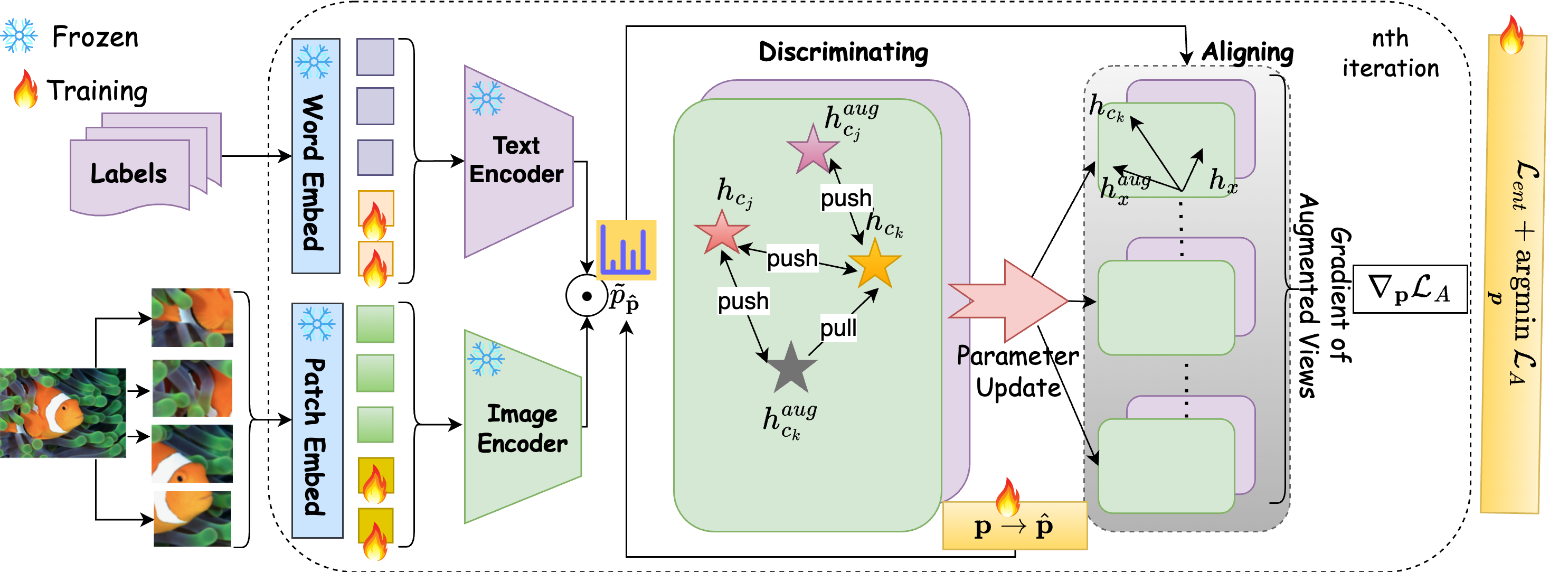}
    \caption{\textbf{Architecture of proposed PromptSync method for zero-shot generalization in CLIP}. During test time, we updates the learnable prompts using discriminative and alignment of class prototypes. For a single test example, we obtain multiple augmented views and obtain the mean class probabilities after parameter updates with discriminating loss. Mean class probabilities act as weights in the class-prototype alignment with filtered augmented views. Gradient are accumulated over multiple iterations before final update to the learnable prompts.}
    \label{fig:arch}
\end{figure*}
\subsection{Proposed Method: PromptSync}
The multi-modal test-time prompt tuning method, PromptAlign \cite{samadh2023align}, updates text and visual prompts using entropy loss and distribution alignment loss with highly confident augmented views (obtained from a test sample $\textbf{X}_{\text{test}}$). PromptAlign, despite considering the distribution, does not take into account the fact that the distribution of each class/domain can be entirely different from other classes/domains, and hence using the source statistics of mean and variance for distribution alignment can still be suboptimal. Inspired by prototype learning \cite{tan2022fedproto} and Extreme-Multi-PatchSSL (EMP-SSL) \cite{tong2023emp}, which establish a prototype/benchmark for each class/sample, we propose class-wise prototype alignment between original and augmented views for both source and test samples. The architecture of PromptSync is shown in Figure \ref{fig:arch}. We use the parameter update from prototype discrimination to generate the class probabilities for the test sample and its augmented views. We accumulate the average of gradients from prototype alignment loss weighted by class probabilities for confident augmented views. The accumulated gradient over multiple iterations is then applied for prompt tuning during test-time adaptation.
\subsection{Class-aware Prototype Generation} 
We generated prototypes for each class for both text and visual branches. The prototype for each class is computed using proxy source dataset. For a test sample, $\textbf{X}_{test}$ and its $N_k$ random views (generated using a set of augmentations $\mathcal{H}$ on $\textbf{X}_{test}$) the prototype vector is generated. Let's denote the token $e_i$ features of a sample $x \in \{\textbf{X}_{test} + \mathcal{H}(\textbf{X}_{test})\}$, at the output of the text encoder and visual encoder as $E_T(x, e_i)$ and $E_V(x, e_i)$, respectively. The prototype for a sample from text and visual branches is given as:
\begin{equation}
    \begin{aligned}
        h_x^{\{t,v\}} &= \frac{1}{|P|}\sum_{i=1}^{|P|} E_{\{T,V\}}(x, e_i) \\
        h_{\textrm{\tiny{CLS}}, x}^v &= E_V(x, e_{\textrm{\tiny{CLS}}})
    \end{aligned}
\end{equation}
where $|P|$ represents the total number of tokens (learnable and non-learnable for both text and visual), excluding EOS, SOS, and CLS. $t, v$ represents textual branch and visual branch respectively. For the proxy source dataset, the class-aware prototype is obtained as:
\begin{align}
        h_{c_k}^t=\frac{1}{|\mathcal{D}(c_k)|}\sum_{x \in \mathcal{D}(c_k)}h_x^t, \\
        h_{c_k}^v=\frac{1}{|\mathcal{D}(c_k)|}\sum_{x \in \mathcal{D}(c_k)}h_x^v, \\ 
        h_{\textrm{\tiny{CLS}}, c_k}^v=\frac{1}{|\mathcal{D}(c_k)|}\sum_{x \in \mathcal{D}(c_k)} h_{\textrm{\tiny{CLS}}, x}^v
\end{align}
where $\mathcal{D}(c_k)$ contains all samples for class $c_k$. The prototypes for augmented views are calculated using the augmented samples for each class $c_k$ denoted as $\mathcal{D}^{aug}(c_k)$ and the corresponding prototypes are denoted as $h_{c_k}^{aug,t}$, $h_{c_k}^{aug, v}$ and $h_{\textrm{\tiny{CLS}}, c_k}^{aug, v}$ respectively.
\subsection{Prototype Discriminating Loss} 
\label{pdl}
The discriminating loss is responsible for training learnable prompts to distinguish the context of samples from one class compared to other classes. This goal is achieved by pushing the class prototype $h_{c_k}^{m}, \textrm{ } m \in \{t, v\}$ for both text and visual branches away from the prototype of class $h_{c_j}^m, \textrm{ } m \in \{t, v\}$ where $c_k \neq c_j$. Likewise, we pull prototypes $h_{c_k}^m$ and $h_{c_k}^{aug,m}$ for same class and push away augmented ones for $c_j \neq c_k$. In this regard, contrastive learning \cite{chen2020simple, he2020momentum, chen2020improved, khosla2020supervised} offers a solution to pull prototypes of positive pairs and push away negative pairs. We refer to \cite{sha2023can} to propose our discriminating loss $\mathcal{L}_{D}$, formally expressed as:
\begin{equation}
    \mathcal{L}_{pos}(c_k) = \frac{1}{|\mathcal{H}|}\sum_{aug \in \mathcal{H}} e^{\textrm{sim}(h_{c_k}^m, h_{c_k}^{aug,m})/\tau}
\end{equation}
\begin{multline}
        \mathcal{L}_{neg}(c_k) = \frac{1}{|\mathcal{H}|}\sum_{aug \in \mathcal{H}} \sum_{c=1, c \neq c_k}^{C}  e^{\textrm{sim}(h_{c_k}^m, h_{c}^m)/\tau} \\ +  e^{\textrm{sim}(h_{c_k}^{aug,m}, h_{c}^m)/\tau} +  e^{\textrm{sim}(h_{c_k}^{m}, h_{c}^{aug,m})/\tau}
\end{multline}
\begin{equation}
    \mathcal{L}_{D} = -\frac{1}{|m| * C} \sum_{\forall m}\sum_{c=1}^{C} \log \frac{\mathcal{L}_{pos}}{\mathcal{L}_{neg}}
\end{equation}
where $c_k \in [1, C]$ and $m \in \{t, v\}, i.e. |m|=2$. The prototypes $h_{c_k}^m$ and $h_{c_k}^{aug,m}$ additionally contains $h_{\textrm{\tiny{CLS}}, c_k}^v, h_{\textrm{\tiny{CLS}}, c_k}^{aug, v}$ when $m=v$. Resulting prompt update $\mathbf{p} \rightarrow \hat{\mathbf{p}}$ (learnable prompt tokens) is obtained after applying gradients for the discriminating loss. Since the proxy dataset will remain same for all the test instances, the updated prompt can be saved and restored each time for an incoming test sample. We presented the study on performance and latency with and without saving these updated prompts in Appendix \ref{ablate:platency}. For the rest of the paper we generalize our method without requiring to save these updated prompts. 
\vspace{-1.4mm}
\subsection{Prototype Alignment Loss} 
\vspace{-1.4mm}
Loss $\mathcal{L}_{D}$ can effectively separate different classes, it is not able to tune the prompt for the test sample, which comes from a different distribution than the source distribution. Hence, we propose the prototype alignment of the test sample (and its augmented views) with the class prototype obtained from the source distribution. We propose to weigh the prototype alignment by the probability of the test sample lying in the particular class. Lets denote the probability $\tilde{p}_{\mathbf{\hat{p}}}[c]$ (as mentioned in Eq.\ref{tpt}) as the mean of probabilities (for class c) produced with the  updated prompt $\hat{\mathbf{p}}$ across filtered augmented views (preserved after the confidence selection filter) including test sample ($F$). The amplitude and angle alignment of sample $x_i$ with the class prototypes for both text and visual branches is calculated as follows:
\begin{align}
    \mathcal{L}_{amp}^{'}(x_i) =  \sum_{c=1}^{C}\tilde{p}_{\mathbf{\hat{p}}}[c] ||p_{x_i}^{m} - p_{c}^{m}||^2 \\
    \mathcal{L}_{ang}^{'}(x_i) = \sum_{c=1}^{C}\tilde{p}_{\mathbf{\hat{p}}}[c] \textrm{sim}(p_{x_i}^{m}, p_{c}^{m})
\end{align}
where $m \in \{t,v\}$. However, there is an issue with MSE loss since it gives an equal penalty (e.g. $\mathcal{L}_{amp}^{'}$ = 0.1) for an increase from 1.2 to 1.3 and 1.7 to 1.8. But we wanted to penalise more for 1.2 to 1.3 since increase in MSE in the smaller range should be penalised more to preserve the base class performance. Hence we penalise with logarithm i.e. we use $\mathcal{L}_{amp} = \log \mathcal{L}_{amp}^{'}$. Similarly, the penalty should be applied to $1/\mathcal{L}_{ang}^{'}$ for angle alignment. The updated amplitude and angle alignment loss is $\mathcal{L}_{amp}(x_i) = \log  \mathcal{L}_{amp}^{'}$ and $\mathcal{L}_{ang}(x_i) =  \log (1/\mathcal{L}_{ang}^{'}) = -\log\mathcal{L}_{ang}^{'}$ respectively. We combined the amplitude and angle loss with equal importance and hence the prototype alignment loss $\mathcal{L}_{A}$ is given as:
\begin{equation}
\begin{aligned}
    \mathcal{L}_{A} = \frac{1}{|F|}\sum_{x_i \in F}(\mathcal{L}_{amp}(x_i) + \mathcal{L}_{ang}(x_i))  \\
    = -\frac{1}{|F|} \sum_{x_i \in F} \log \frac{\mathcal{L}_{ang}'(x_i)}{\mathcal{L}_{amp}'(x_i)}
\end{aligned}
\label{la}
\end{equation}
\subsection{Algorithm Details} 
In order to compute the prototype discriminating loss on the source dataset, we require the pre-training dataset of the CLIP model. However, it was trained on over 400 million image-text pairs, which are not publicly available. Nevertheless, in previous works\cite{bahng2022exploring, samadh2023align}, CLIP has been heavily tuned on the ImageNet\cite{deng2009imagenet} dataset to achieve excellent zero-shot performance. Hence, we use ImageNet as the proxy for the source dataset to compute prototypes for each class. These prototypes are computed offline for both the sample and its augmented views, and they are used directly during test-time adaptation. During each iteration of test-time adaptation, the meta-train stage is entered first. The model starts training using the prototype discriminating objective $\underset{\boldsymbol{p}}{\textrm{\textbf{argmin} }} \mathcal{L}_{D}$, and gradients are calculated, resulting in the prompt update $\mathbf{p} \rightarrow \hat{\mathbf{p}}$ ($i^{th}$ iteration). Subsequently, the meta-test stage is executed. Here, the augmented views are first filtered using a confidence threshold over predicted probabilities using the updated prompts $\hat{\mathbf{p}}$. The mean probabilities $\tilde{p}_{\mathbf{\hat{p}}}$ are computed over $F$ and used as weights in $\mathcal{L}_{A}$. The model is trained on $F$, and the gradient of prototype alignment loss $\nabla_{{\mathbf{p}}}\mathcal{L}_{A}$ is calculated. We average out the gradients over all samples in $F$. Finally, the prompts $\boldsymbol{p}$ is updated using combined objective: $\mathcal{L}_{ent} +  \underset{\boldsymbol{p}}{\textrm{\textbf{argmin} }} \mathcal{L}_{A}$. For $n > 1$ we accumulate the averaged gradients before final prompt update.
\begin{table*}[h]
\centering
\small
\begin{tabular}{lccccc}
\toprule
            & Imagenet V2 & Imagenet Sketch & Imagenet A & Imagenet R & OOD Avg \\ \midrule
CLIP        &    60.86          & 46.09                 &   47.87          &     73.98        &   57.20      \\
CLIP+TPT    &     64.35          &        47.94         &   54.77           &     77.06       &   60.81       \\
CoOp        &    64.20          &        47.99          &   49.71          &    75.21         &    59.28     \\
CoOp+TPT    &   66.83           &      49.29            &   57.95          &   77.27         &    62.84      \\
Co-CoOp     &  64.07            &        48.75          & 50.63           &     76.18        & 59.91         \\
Co-CoOp+TPT &     64.85          &        48.27          &   58.47         &     78.65         &   62.61      \\
MaPLe &     64.07         & 49.15                 &    50.90         &    76.98        &   60.28       \\
MaPLe+TPT  &   64.87            &       48.16           &  58.08          &   78.12          &   62.31       \\
PromptAlign &   65.29           &         50.23         &    59.37        &     79.33         &    63.55     \\ \midrule
\rowcolor[HTML]{C0C0C0}
PromptSync  &     \textbf{67.54}       &   \textbf{53.42}              &       \textbf{61.92}     &     \textbf{80.64}       &  \textbf{65.88 }      \\ \bottomrule
\end{tabular}
\caption{\textbf{Comparison on the domain generalization setting}. Prompt tuning methods
are trained on ImageNet and evaluated on datasets with domain shifts}
\label{tab:domainG}
\end{table*}
\begin{table*}
\centering
\small
\begin{tabular}{lllllll}
\toprule
\multicolumn{1}{c}{} & \multicolumn{1}{c}{\begin{tabular}[c]{@{}c@{}}Camera\\ (Yaw/ Pitch/ Roll)\end{tabular}} & \multicolumn{1}{c}{\begin{tabular}[c]{@{}c@{}}Pose\\ (Yaw/ Pitch/ Roll)\end{tabular}} & \multicolumn{1}{c}{Scale} & \multicolumn{1}{c}{Texture} & \multicolumn{1}{c}{Lighting} & \multicolumn{1}{c}{Worlds} \\ \midrule
MaPLe                &     48.73/ 39.93/ 32.13                                                                                     &   48.10/ 28.40/ 27.80                                                                                     &    46.90                         &    37.90                          &   15.50                           &    32.13                         \\
MaPLe+TPT            &   57.04/ 45.99/ 39.23                                                                                     &   56.26/ 35.64/ 33.26                                                                                      &   54.87                          &       43.73                       &         22.52                    &    42.00                          \\
PromptAlign          & 58.14/ 46.93/ 40.45                                                                                         &  57.43/ 36.31/ 34.32                                                                                     &    56.18                         &     44.97                       &      23.06                       &    43.24                            \\ \midrule
\rowcolor[HTML]{C0C0C0} 
PromptSync           &  \textbf{59.84/ 48.54/ 41.92}                                                                                       &  \textbf{59.72/ 38.84/ 36.64}                                                                                   &     \textbf{58.12}                         &     \textbf{45.98}                      &      \textbf{25.02}                       &    \textbf{44.84}     \\ \midrule
\end{tabular}
\caption{\textbf{Comparison on the domain generalization setting for distribution alignment}. MaPLe is trained on ImageNet and evaluated on OOD dataset i.e., PUG}
\label{tab:domainpug}
\end{table*}
\section{Experiments}
\label{sec:exp}
We have evaluated PromptSync on different benchmark settings (Appendix \ref{sec:benchmark}) with different datasets described below:\\
\textbf{Datasets}: For domain generalisation setting, we follow PromptAlign \cite{samadh2023align} and evaluated our method on four out-of-distribution (OOD) variants of ImageNet \cite{deng2009imagenet}: ImageNetV2 \cite{recht2019imagenet}, ImageNet-Sketch \cite{wang2019learning}, ImageNet-A \cite{hendrycks2021natural} and ImageNet-R \cite{hendrycks2021many}. We also consider the evaluation on a recent and challenging benchmark, namely, Photorealistic Unreal Graphics (PUG) dataset \cite{bordes2023pug}, comprised of different textures, sizes, orientations, and backgrounds. For cross-dataset transfer setting, we follow TPT \cite{shu2022test} and evaluate the performance on 10 diverse image classification datasets with varying complexities for visual recognition tasks. This includes Caltech 101 \cite{fei2004learning} for generic objects. Five fine-grained datasets (spanning images of animals, flowers and transportation) are StanfordCars \cite{krause20133d}, Food101 \cite{bossard2014food}, Flowers102 \cite{nilsback2008automated}, FGVC-Aircraft \cite{maji2013fine}, OxfordPets \cite{parkhi2012cats}. Moreover, four datasets, namely, SUN397 \cite{sun2020test}, DTD \cite{cimpoi2014describing}, UCF101 \cite{soomro2012dataset}, and EUROSAT \cite{helber2019eurosat}, comprise scenes, textures, human actions, and satellite imagery, respectively. For base-to-novel generalisation, we follow \cite{khattak2023maple} and evaluate our method on ImageNet and the 10 image classification datasets.\\
\textbf{Baselines}: We compared PromptSync with existing few-shot prompt learning methods for CLIP adaptation; these are CoOp \cite{zhou2022learning}, CoCoOp \cite{zhou2022conditional}, TPT \cite{shu2022test}, and PromptAlign \cite{samadh2023align}. MaPLe \cite{khattak2023maple} is a multi-modal prompt learning baseline that adapts CLIP by learning prompts on both text and visual branches. TPT \cite{shu2022test} and PromptAlign \cite{samadh2023align} are the test-time prompt tuning methods that tune the prompt for each incoming test sample, achieving state-of-the-art performance in prompt learning.\\
\textbf{Implementation Details}: We ran all experiments on a single NVIDIA A100 40GB GPU. Following \cite{khattak2023maple}, we trained on ImageNet with 16-shot training data selected at random for each class using 2 prompt tokens for a depth of 3 layers (on CLIP ViT-B/16 backbone architecture). We optimized the prompts on both the text and visual branches using a single test image. We augmented each test image with 127 different views using random resized crops, background substitution, horizontal flip augmentations, and visual corruption. For text augmentation, we used hyponyms, synonyms, and meronyms from WordNet\cite{miller1995wordnet}. Moreover, we generated various text prompts from pre-trained LLMs \cite{brown2020language}. Additionally, we randomly masked one of the learnable tokens for 15\% of augmented views. We computed the gradients of alignment loss for a batch size of 128 images, including the original image. During the meta-train stage, we updated the original parameters (using a single iteration) and then optimized the prompts in the meta-test stage by calculating the gradients of alignment loss w.r.t. the updated parameters accumulated for a single ($n=1$) iteration to facilitate the one-to-one comparison with baselines. We obtained the top 10\% confident predictions of augmented views based on the lowest entropy. We used the AdamW optimizer and a learning rate $\beta$ of $5e^{-4}$ for the fine-grained datasets and $0.04$ for the rest of the datasets.
\subsection{Domain Generalization}
We demonstrate that all test-time adaptation methods exhibit better performance (Table \ref{tab:domainG}) compared to the pre-trained CLIP model, highlighting the advantage of tuning V-L models at test time. PromptSync achieves the highest Top-1 accuracy averaged across all the domains of ImageNet variants. Furthermore, we evaluated the ImageNet-trained model on various out-of-distribution (OOD) datasets and observed consistent improvement in performance compared to existing state-of-the-art (SOTA) approaches. The detailed results for each domain dataset are presented in Tables \ref{tab:domainG} and \ref{tab:domainpug}. This confirms that alignment and discriminative training with augmented views on both the text and visual branches enhance the generalization performance of V-L models like CLIP.
\begin{table*}[]
\centering
\small
\resizebox{\textwidth}{!}{%
\begin{tabular}{ll|c|c|c|c|c|c|c}
\toprule
\rowcolor[HTML]{C0C0C0} 
\multicolumn{1}{c}{Datasets} & \multicolumn{1}{c}{Sets}                             & \multicolumn{1}{c}{\begin{tabular}[c]{@{}c@{}}CoOp\\ (IJCV22)\end{tabular}} & \multicolumn{1}{c}{\begin{tabular}[c]{@{}c@{}}CoCoOp\\ (CVPR22)\end{tabular}}  & \multicolumn{1}{c}{\begin{tabular}[c]{@{}c@{}}ProDA\\ (CVPR22)\end{tabular}} &  \multicolumn{1}{c}{\begin{tabular}[c]{@{}c@{}}MaPLe\\ (CVPR23)\end{tabular}} & \multicolumn{1}{c}{\begin{tabular}[c]{@{}c@{}}MaPLe + TPT\\ (CVPR23)\end{tabular}} & \multicolumn{1}{c}{\begin{tabular}[c]{@{}c@{}}PromptAlign\\ (NIPS23)\end{tabular}} & \multicolumn{1}{c}{\begin{tabular}[c]{@{}c@{}}PromptSync\end{tabular}}\\ \midrule
Average                      & \begin{tabular}[c]{@{}l@{}}Base\\ Novel\end{tabular} & \begin{tabular}[c]{@{}l@{}}82.38\\ 67.96 \end{tabular}                                                                             &  \begin{tabular}[c]{@{}l@{}}80.47  \\ 71.69\end{tabular}                                                                            &   \begin{tabular}[c]{@{}l@{}} 81.56\\ 72.30 \end{tabular}                                                                          &    \begin{tabular}[c]{@{}l@{}} 82.24 \\ 75.09\end{tabular}                                                                             &   \begin{tabular}[c]{@{}l@{}} 82.16 \\ 74.95 \end{tabular}                                                                          &   \begin{tabular}[c]{@{}l@{}}83.19  \\ 75.88\end{tabular}                                                                               &   \begin{tabular}[c]{@{}l@{}}{84.17}  \\ {77.17}\end{tabular}                                 \\ \midrule
ImageNet                     & \begin{tabular}[c]{@{}l@{}}Base\\ Novel\end{tabular} &  \begin{tabular}[c]{@{}l@{}}76.46\\ 66.31 \end{tabular}                                                                            &    \begin{tabular}[c]{@{}l@{}}75.98\\ 70.43\end{tabular}                                                                            &  \begin{tabular}[c]{@{}l@{}}75.40\\ 70.23\end{tabular}                                                                            &  \begin{tabular}[c]{@{}l@{}}76.67\\ 70.54 \end{tabular}                                                                                &    \begin{tabular}[c]{@{}l@{}}77.73\\ 72.24 \end{tabular}                                                                            &     \begin{tabular}[c]{@{}l@{}}78.26\\ 72.59\end{tabular}                                                                          &  \begin{tabular}[c]{@{}l@{}}79.23\\ 73.84\end{tabular}                                                                                                                                       \\ \midrule
Caltech101                   & \begin{tabular}[c]{@{}l@{}}Base\\ Novel\end{tabular} &   \begin{tabular}[c]{@{}l@{}}97.80\\ 93.27 \end{tabular}                                                                          &                                \begin{tabular}[c]{@{}l@{}}97.96\\ 93.81\end{tabular}                                               &  \begin{tabular}[c]{@{}l@{}} 98.27\\ 93.23 \end{tabular}                                                                         &  \begin{tabular}[c]{@{}l@{}}98.00\\ 94.27\end{tabular}                                                                                 &  \begin{tabular}[c]{@{}l@{}}98.54\\  94.29 \end{tabular}                                                                            &  \begin{tabular}[c]{@{}l@{}}98.60\\ 94.50\end{tabular}                                                                               &    \begin{tabular}[c]{@{}l@{}}98.62\\ 94.67\end{tabular}                                                                                                          \\ \midrule
OxfordPets                   & \begin{tabular}[c]{@{}l@{}}Base\\ Novel\end{tabular} &     \begin{tabular}[c]{@{}l@{}}94.47 \\ 96.00 \end{tabular}                                                                        &   \begin{tabular}[c]{@{}l@{}}95.20\\ 97.69\end{tabular}                                                                            & \begin{tabular}[c]{@{}l@{}}95.43\\ 97.83 \end{tabular}                                                                                &    \begin{tabular}[c]{@{}l@{}}95.43\\ 97.80 \end{tabular}                                                                            &     \begin{tabular}[c]{@{}l@{}}95.23\\ 97.37 \end{tabular}                                                                        &    \begin{tabular}[c]{@{}l@{}}95.38\\ 97.56\end{tabular}                                                                                &    \begin{tabular}[c]{@{}l@{}}95.44\\ 97.83\end{tabular}                                                                     \\ \midrule
Stanford Cars                & \begin{tabular}[c]{@{}l@{}}Base\\ Novel\end{tabular} &   \begin{tabular}[c]{@{}l@{}}75.67\\ 67.53 \end{tabular}                                                                          &                          \begin{tabular}[c]{@{}l@{}} 70.49\\ 73.59\end{tabular}                                                     &   \begin{tabular}[c]{@{}l@{}}74.70\\ 71.20 \end{tabular}                                                                          & \begin{tabular}[c]{@{}l@{}}72.90\\ 73.97 \end{tabular}                                                                                &     \begin{tabular}[c]{@{}l@{}}74.00\\ 75.20 \end{tabular}                                                                         &  \begin{tabular}[c]{@{}l@{}}75.02\\ 75.71\end{tabular}                                                                             &    \begin{tabular}[c]{@{}l@{}}76.42\\ 77.21\end{tabular}                                                                                                  \\ \midrule
Flowers102                   & \begin{tabular}[c]{@{}l@{}}Base\\ Novel\end{tabular} &                    \begin{tabular}[c]{@{}l@{}}97.27 \\ 67.13 \end{tabular}                                                         &    \begin{tabular}[c]{@{}l@{}}94.87\\ 71.75\end{tabular}                                                                           &    \begin{tabular}[c]{@{}l@{}}97.70 \\ 68.68 \end{tabular}                                                                         &   \begin{tabular}[c]{@{}l@{}}95.93 \\ 72.40 \end{tabular}                                                                             &     \begin{tabular}[c]{@{}l@{}}96.24 \\ 72.10 \end{tabular}                                                                         &     \begin{tabular}[c]{@{}l@{}}96.61\\ 72.34\end{tabular}                                                                          &    \begin{tabular}[c]{@{}l@{}}97.73\\ 73.78\end{tabular}                                                                                        \\ \midrule
Food101                      & \begin{tabular}[c]{@{}l@{}}Base\\ Novel\end{tabular} &                                 \begin{tabular}[c]{@{}l@{}}89.37 \\ 88.77 \end{tabular}                                            &                                         \begin{tabular}[c]{@{}l@{}}90.70\\ 91.29\end{tabular}                                      &   \begin{tabular}[c]{@{}l@{}}90.30 \\ 88.57 \end{tabular}                                                                          &    \begin{tabular}[c]{@{}l@{}}90.70 \\ 92.07 \end{tabular}                                                                            &    \begin{tabular}[c]{@{}l@{}}91.13 \\ 92.03 \end{tabular}                                                                          &      \begin{tabular}[c]{@{}l@{}}91.63\\ 92.68\end{tabular}                                                                         &   \begin{tabular}[c]{@{}l@{}}92.39\\ 92.95\end{tabular}                                                                                       \\ \midrule
FGVC Aircraft                & \begin{tabular}[c]{@{}l@{}}Base\\ Novel\end{tabular} &                                    \begin{tabular}[c]{@{}l@{}}39.67 \\ 31.23 \end{tabular}                                         &                                              \begin{tabular}[c]{@{}l@{}}33.41\\23.71\end{tabular}                                 &    \begin{tabular}[c]{@{}l@{}}36.90 \\ 34.13 \end{tabular}                                                                         &    \begin{tabular}[c]{@{}l@{}}37.27 \\ 35.53 \end{tabular}                                                                            &   \begin{tabular}[c]{@{}l@{}}34.31 \\\ 35.81 \end{tabular}                                                                           &    \begin{tabular}[c]{@{}l@{}}37.21\\ 37.27\end{tabular}                                                                           &     \begin{tabular}[c]{@{}l@{}}40.91\\ 39.31\end{tabular}                                                                                        \\ \midrule
SUN397                       & \begin{tabular}[c]{@{}l@{}}Base\\ Novel\end{tabular} &                                \begin{tabular}[c]{@{}l@{}}80.85 \\ 68.34 \end{tabular}                                             &   \begin{tabular}[c]{@{}l@{}}79.74\\ 76.86\end{tabular}                                                                            &                            \begin{tabular}[c]{@{}l@{}}78.67 \\ 76.93 \end{tabular}                                                 &                                                  \begin{tabular}[c]{@{}l@{}}80.80 \\ 78.70 \end{tabular}                              &        \begin{tabular}[c]{@{}l@{}}81.15 \\ 79.18 \end{tabular}                                                                      &                                        \begin{tabular}[c]{@{}l@{}}81.57\\ 79.48\end{tabular}                                       &   \begin{tabular}[c]{@{}l@{}}84.28\\ 83.01\end{tabular}                                                                                                   \\ \midrule
DTD                          & \begin{tabular}[c]{@{}l@{}}Base\\ Novel\end{tabular} &                                  \begin{tabular}[c]{@{}l@{}}79.97 \\ 48.60 \end{tabular}                                           &                                            \begin{tabular}[c]{@{}l@{}}77.01\\ 56.00\end{tabular}                                   &                                                 \begin{tabular}[c]{@{}l@{}}80.67 \\ 56.48 \end{tabular}                            &                                                    \begin{tabular}[c]{@{}l@{}}80.30 \\ 59.23 \end{tabular}                            &                                                    \begin{tabular}[c]{@{}l@{}}82.20 \\ 59.91 \end{tabular}                          &     \begin{tabular}[c]{@{}l@{}}82.60\\ 60.55\end{tabular}                                                                          &    \begin{tabular}[c]{@{}l@{}}83.49\\ 62.03\end{tabular}                                                                                                   \\ \midrule
Eurosat                      & \begin{tabular}[c]{@{}l@{}}Base\\ Novel\end{tabular} &                                   \begin{tabular}[c]{@{}l@{}}90.10 \\ 53.00 \end{tabular}                                          &                                            \begin{tabular}[c]{@{}l@{}}87.49\\ 60.04\end{tabular}                                   &                                               \begin{tabular}[c]{@{}l@{}}83.90 \\ 66.00\end{tabular}                              &                                                  \begin{tabular}[c]{@{}l@{}}93.63 \\  72.87 \end{tabular}                              &                                              \begin{tabular}[c]{@{}l@{}}91.02 \\ 68.96 \end{tabular}                                &                                                  \begin{tabular}[c]{@{}l@{}}94.10\\ 72.71\end{tabular}                             &                                                   \begin{tabular}[c]{@{}l@{}}94.63\\ 73.19\end{tabular}                                     \\ \midrule
UCF101                       & \begin{tabular}[c]{@{}l@{}}Base\\ Novel\end{tabular} &                              \begin{tabular}[c]{@{}l@{}}84.53 \\ 67.37 \end{tabular}                                               &                                    \begin{tabular}[c]{@{}l@{}}82.33\\ 73.45\end{tabular}                                           &                                     \begin{tabular}[c]{@{}l@{}}85.23 \\ 71.97 \end{tabular}                                        &                                        \begin{tabular}[c]{@{}l@{}}82.97 \\ 78.57 \end{tabular}                                        &                                         \begin{tabular}[c]{@{}l@{}}82.23 \\ 77.34\end{tabular}                                     &                                       \begin{tabular}[c]{@{}l@{}}84.11\\  79.30\end{tabular}                                        &                                       \begin{tabular}[c]{@{}l@{}}85.75\\ 81.29\end{tabular}                                                                    \\ \bottomrule
\end{tabular}%
}
\caption{\textbf{Comparison on Base-to-novel generalization setting}. PromptSync shows consistent improvement on both base and novel classes over previous methods}
\label{tab:basenovel}
\end{table*}
\subsection{Base to Novel Generalization}
Table \ref{tab:basenovel} presents the detailed performance report of PromptSync on base and novel classes across 11 recognition datasets. On average, our strategy outperforms the model performance by 1.29\% on base classes and nearly 1\% on novel classes. We observe that PromptAlign, based on a distribution alignment strategy, outperforms for novel classes in most cases, with an average improvement of 0.79\% compared to the best-performing model. However, the margin of improvement is very low. In contrast, with TPT, the performance drops in some instances, such as for OxfordPets, Eurosat, and UCF101. This demonstrates that: 1) test-source alignment is crucial for prompt tuning. 2) Prompt tuning alone in the text branch is not sufficient for zero-shot generalization. Since distribution alignment does not promote discriminative learning and the entropy loss on the test dataset is noisy, PromptSync outperforms with class-aware prototype discrimination and alignment across different augmented views. Averaging the gradients further motivates domain-agnostic prompt tuning on both the text and visual branches. This enhances the zero-shot generalization of the V-L model compared to other state-of-the-art approaches. Moreover, our strategy for prompt tuning does not lose information for base classes.
\begin{table*}
\centering
\resizebox{\textwidth}{!}{%
\begin{tabular}{llllllllllll}
\toprule
            & Caltech & Pets & Cars & Flowers & Food101 & Aircraft & SUN397 & DTD & EuroSAT & UCF101 & Average \\ \midrule
CLIP        &   93.35       &   88.25    &   65.48    &   67.44       &    83.65      &    23.67       &   62.59      &  44.27    &  42.01        &    65.13     & 63.58        \\
CLIP+TPT    &  94.16       &      87.79  &  66.87     &  68.98        &    84.67     &     24.78      &   65.50       &  47.75   &  42.44         &   68.04      &  65.10       \\
CoOp        &   93.70      &  89.14      &  64.51     & 68.71         &   85.30      &    18.47        &   64.15      &  41.92    &      46.39    &  66.55       &   63.88      \\
CoOp + TPT       &   93.15       &  89.48     &   66.77     &  68.48        &   86.48       &  20.51         &   66.06      &  43.32   &       37.73    &   68.91     &  64.08        \\
CoCoOp      &   93.79     &  90.46     &  64.90       & 70.85          &   83.97        &    22.29     &  66.89   &  45.45         &   39.23      & 68.44 & 64.63        \\
CoCoOp + TPT     &  88.57       &   85.33    & 59.68      & 55.31          &   80.64       &  16.89         &  60.24      &  38.93    &       48.55    &   63.35      &  59.75       \\
ProDA       &  86.70        &  88.20     &   60.10    &   77.50       &   80.80      &   22.20         &    -     &   50.90   &  58.50        &    -     &     65.62    \\
MaPLe       &    93.53      &  90.49     &   65.57    &  72.23       &  86.20        &   24.74         &    67.01     &  46.49    &      48.06    &    68.69     &   66.30      \\
MaPLe+TPT   &   93.59       &   90.72    &   66.50    &  72.37        &  86.64        &  24.70         &   67.54      & 45.87     &       47.80   &      69.19   &   66.50      \\
PromptAlign &   94.01      &   90.76     &  68.50   &    72.39        &    86.65      &  24.80         &    67.54     &   47.24   &       47.86   &  69.47      &    66.92      \\ \midrule
\rowcolor[HTML]{C0C0C0} 
PromptSync  &   \textbf{95.78}      &  \textbf{91.89}    &  \textbf{69.24}    &  \textbf{77.68}        &  \textbf{87.72}      &   \textbf{25.91}       &  \textbf{67.98}       &  \textbf{50.99}   &   \textbf{59.36}      &  \textbf{71.04}       &  \textbf{69.76}       \\ \bottomrule
\end{tabular}%
}
\caption{\textbf{Comparison on cross-dataset transfer setting}. Prompt tuning methods are
trained on ImageNet and evaluated on cross-datasets}
\label{tab:crossdatasets}
\end{table*}
\begin{table}[]
\centering
\resizebox{\columnwidth}{!}{%
\begin{tabular}{lcccc}
\toprule
\multicolumn{1}{c}{Method} & \multicolumn{1}{c}{\begin{tabular}[c]{@{}c@{}}Entropy\\ Loss\end{tabular}} & \multicolumn{1}{c}{\begin{tabular}[c]{@{}c@{}}Alignment\\ Loss\end{tabular}} & \multicolumn{1}{c}{\begin{tabular}[c]{@{}c@{}}Discriminative\\ Loss\end{tabular}} & \multicolumn{1}{c}{\begin{tabular}[c]{@{}c@{}}Top-1 \\ Acc.\end{tabular}} \\ \midrule
MaPLe                      &      \xmark                                                                       &     \xmark                                                                         &                                                              \xmark                     & 50.90                                                                         \\
MaPLe+TPT                  &      $\checkmark$                                                                      &    \xmark                                                                         &  \xmark                                                                                &    58.08                                                                       \\
PromptAlign$^{\dagger}$                &   \xmark                                                                        &   $\checkmark$                                                                          &    \xmark                                                                              &   50.85                                                                        \\
PromptAlign                &    $\checkmark$                                                                       &     $\checkmark$                                                                        &     \xmark                                                                            &    59.37                                                                       \\ 
PromptSync${^\dagger}$                 &      \xmark                                                                    &      \xmark                                                                      &          $\checkmark$                                                                        &    56.67                                                                       \\
\midrule
\rowcolor[HTML]{C0C0C0} 
PromptSync                 &      $\checkmark$                                                                     &       $\checkmark$                                                                        &          $\checkmark$                                                                        &       \textbf{61.92}                                                                    \\ \bottomrule
\end{tabular}%
}
\caption{\textbf{Ablation Study}. Analysis of Alignment, Discriminative and Entropy minimization loss. The average of Top-1 accuracy(\%) across three seeds is reported}
\label{tab:ablation}
\end{table}

\subsection{Cross-Dataset Transfer}
In Table \ref{tab:crossdatasets}, we compared the transfer performance of PromptSync with existing state-of-the-art methods using prompt learning. We evaluated methods for transfer performance across diverse cross-datasets. PromptSync consistently outperforms the previous best method, i.e., PromptAlign \cite{samadh2023align}, across all cross-datasets, providing an average improvement of $2.84\%$. Compared to PromptAlign, which outperforms the previous method MaPLe + TPT by a very small margin, i.e., $0.42\%$, our method shows a significant average improvement of $3.26\%$ over MaPLe + TPT. Other methods, CoOp and CoCoOp, on average, perform worse than zero-shot CLIP + TPT (except ProDA \cite{zhang2021prototypical}). This affirms that both text-visual alignment and domain-agnostic parameter updates result in better transfer generalization across cross-datasets in V-L models. As opposed to our method, the previous approaches were not consistent in performance across all datasets, which further affirms the advantage of a domain-agnostic training strategy.
\begin{table}
\centering
\resizebox{\columnwidth}{!}{%
\begin{tabular}{lccccc}
\toprule
           & $\mathcal{L}_{amp}$ & $\mathcal{L}_{ang}$ & sub & sum\_exp & $\mathcal{L}_{A}$ \\ \midrule
PromptSync &      59.84                                 &                58.81                       &   57.86  &    59.83      &    \textbf{61.92}                                 \\ \bottomrule
\end{tabular}%
}
\caption{\textbf{Ablation analysis to alignment loss variants}. All results are on ImageNet-A dataset}
\label{tab:loss_variant}
\end{table}
\section{Ablation}
\textbf{Class-Aware Prototype Alignment}: Table \ref{tab:ablation} summarizes the comparison between two alignment strategies: distribution alignment of the test sample with the class-agnostic source distribution. All results are on the ImageNet-A dataset. PromptAlign adopted distribution alignment along with averaged cross-entropy for prompt tuning. However, we perform domain-agnostic parameter updates with class-aware prototype alignment for the test sample. As shown in Table \ref{tab:ablation}, PromptAlign$^\dagger$ without entropy loss is as good as vanilla MaPLe. This is due to the fact that distributional alignment does not promote any discriminative learning in the absence of entropy loss. However, because entropy loss is noisy due to the poor performance of the vanilla zero-shot V-L model, we propose the stronger discriminative loss of class prototype alignment for prompt tuning with source and test samples with augmented views. PromptSync$^\dagger$ without entropy loss outperforms the corresponding counterpart PromptAlign$^\dagger$. This is because the class-aware prototype alignment has both alignment and discriminative properties, thus improving test-time adaptation on its own. With additional signals from predicted probabilities for each class, the class-aware prototype alignment acts as a geometric regularizer, mitigating class collapse in prompt representation.\\
\textbf{Loss variants}: We conducted an ablation study on amplitude and angle loss for the class-aware prototype alignment objective. Table \ref{tab:loss_variant} compares three loss choices: 1) amplitude loss, 2) angle loss, and 3) amplitude + angle loss. Clearly, the combination of amplitude and angle performs better than other choices. The formulation for the combination of amplitude and angle loss is the same as in equation \ref{la}. We further investigated other variants, i.e., combining two of them without taking the log: 1) subtraction between amplitude and angle $\mathcal{L}_{amp}^{'} - \mathcal{L}_{ang}^{'}$(\textit{sub}) 2) the summation of exponential of both losses $\textrm{exp}(\mathcal{L}_{amp}^{'}) + \textrm{exp}(1/\mathcal{L}_{ang}^{'})$(\textit{sum\_exp}). Clearly, the formulation in equation \ref{la} ($\mathcal{L}_{A}$) performs best among other variants. Ablation on the proxy dataset is given in Appendix \ref{laion}, and ablation on performance and latency with and without saving updated prompts is provided in Appendix \ref{ablate:platency}. We also compared the number of augmented views and prompt updates in Appendix \ref{sensitivity}.

\section{Performance and Latency}
\label{ablate:platency}
The experiments presented in the Table \ref{tab:platency} (Appendix) involve a comparison of different methods, namely MaPLe + TPT, PromptAlign, PromptSync*, and PromptSync. In these experiments, we evaluated the top-1 average accuracy (\%) and latency (in hours for a single prompt update) of each method. Specifically, we investigated PromptSync with and without saving the updated prompt obtained after prototype discrimination, with the variant denoted as PromptSync* indicating the adaptation of prompt tokens for test samples after restoring saved prompt tokens.

The results, as shown in Table \ref{tab:platency}, include latency measurements represented in hours for a single prompt update, and all evaluations are conducted on the ImageNet-A dataset. Notably, the PromptSync* variant demonstrates a faster processing time compared to the full PromptSync method, with only a marginal drop in performance. This outcome underscores the achieved generalization through prototype alignment. Furthermore, in comparison to previous methods such as MaPLe + TPT and PromptAlign, the PromptSync* variant exhibits only a slight increase in latency (0.03 hours) while still improving overall performance.
\section{Sensitivity Comparison}
\label{sensitivity}
We further performed the sensitivity comparison of our method as compared to other state-of-the-art baselines. In Appendix, Figure \ref{fig:tradeoff}(a) shows the comparison of performance during test time adaptation as the number of views increases. All the
results are on ImageNet-A dataset. In comparison to PromptAlign and MaPLe + TPT, their performance almost plateaus around 64 views with insignificant improvement further, while PromptSync shows a consistent improvement with the increase in views and insignificant improvement beyond 128. This proves the generalizability achieved by our method since it optimises base CLIP over a larger number of possible shifts in the dataset, resulting in better performance. Figure \ref{fig:tradeoff}(b) shows the performance comparison as the number of prompt update steps increases. All the methods increase their performance with an increase in the number of steps; however, our method shows better adaptation to the test sample with more steps in comparison to PromptAlign and MaPLe + TPT. For apples-to-apples comparison we perform a single-step update (128 views) following TPT \cite{shu2022test}. 
\section{LAION400M Proxy Dataset Analysis}
\label{laion}
Given CLIP's impressive zero-shot performance on ImageNet, we opted for ImageNet as a viable proxy source dataset, aligning with prior research \cite{samadh2023align}. We worked with a subset of LAION400M, comprising 2.5 million images (2 times the size of ImageNet). Furthermore, we carried out an ablation study on the alignment strategy using LAION400M as the source dataset, a dataset known to mirror CLIP's training dataset \cite{cherti2023reproducible}. The results for this ablation study is shown in Table \ref{tab:laion} (Appendix). Notably, the performance impact remains consistent when utilizing this subset of LAION400M alongside ImageNet. Source class prototypes are computed on the proxy source data to derive the distribution for alignment during test time. As this proxy dataset aligns with the model's training set, this offline computation remains unchanged despite environmental shifts and only necessitates computation once. 
\section*{Conclusion}
In summary, PromptSync significantly improves zero-shot generalization in vision-language models. Our approach, addressing class dominance and variance, outperforms existing methods by 2.33\% overall, with a 1\% boost in base-to-novel generalization and 2.84\% in cross-dataset transfer on a domain generalization benchmark. This underscores PromptSync's effectiveness in enhancing the robustness of vision-language models.

{
    \small
    \bibliographystyle{ieeenat_fullname}
    \bibliography{main}
}

\clearpage
\setcounter{page}{1}
\maketitlesupplementary
\section{Benchmark Settings}
\label{sec:benchmark}
\begin{table}[]
\centering
\resizebox{\columnwidth}{!}{%
\begin{tabular}{lcc}
\toprule
Method      & Top-1 Average Accuracy(\%) & Latency \\ \midrule
MaPLe + TPT & 58.08                      &  0.41       \\
PromptAlign & 59.37                      &  0.46       \\
PromptSync* & 61.88                      &  0.49       \\ \midrule
PromptSync  & 61.92                      &  0.65       \\ \bottomrule
\end{tabular}%
}
\caption{\textbf{Performance and Latency}: Performance and Latency comparison of PromptSync with state-of-the-art baselines and its variant which reuse the learned prompt tokens after prototype discrimination without learning them for each incoming test sample.}
\label{tab:platency}
\end{table}
\begin{figure*}[h]
\centering
\begin{minipage}[b]{.4\textwidth}
\includegraphics[width=\linewidth]{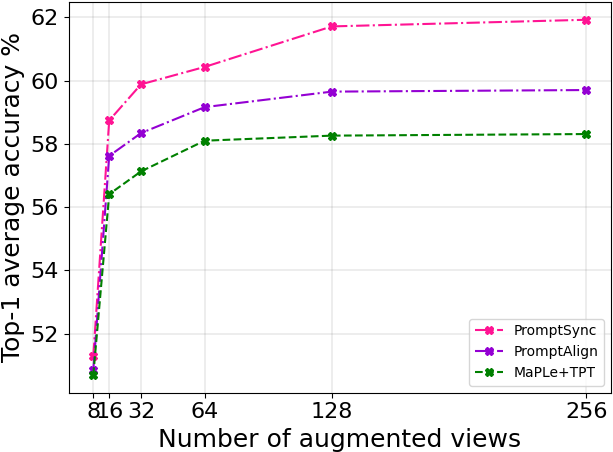}
\end{minipage}\qquad
\begin{minipage}[b]{.4\textwidth}
\includegraphics[width=\linewidth]{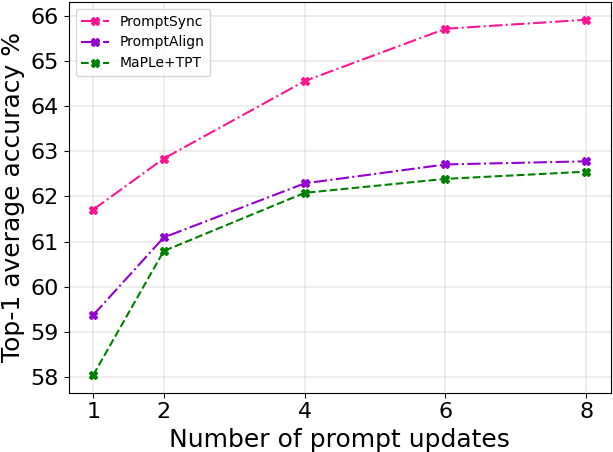}
\end{minipage}
\caption{\textbf{Sensitivity Comparison}. (a) Top-1 accuracy improves with number of augmented views (b) Top-1 accuracy improves consistently with number of prompt update steps.}
\label{fig:tradeoff}
\end{figure*}
\begin{table*}[h]
\centering
\small
\resizebox{\textwidth}{!}{%
\begin{tabular}{llllllllllll}
\toprule
Method   & Flowers & DTD   & Pets  & Cars  & UCF   & Caltech & Food  & SUN   & Aircraft & Eurosat & Avg   \\ \midrule
ImageNet & 77.68   & 50.99 & 91.89 & 69.24 & 71.04 & 95.78   & 87.72 & 67.98 & 25.91    & 59.36   & 69.74 \\
LAION    & 77.68   & 51.00 & 91.88 & 69.25 & 71.03 & 95.79   & 87.75 & 68.00 & 25.90    & 59.35   & 69.76 \\ \bottomrule
\end{tabular}%
}
\caption{Performance impact analysis using both ImageNet and LAION400M subset}
\label{tab:laion}
\end{table*}

\textbf{Base-to-Novel Generalisation}: Following MaPLe \cite{khattak2023maple}, we evaluate PromptSync on a zero-shot setting. We split the dataset into base and novel classes. The model is trained only on the base classes in a few-shot setting and evaluated on the base and novel classes.\\ 
\textbf{Cross-dataset Transfer}: We evaluate PromptSync on the ImageNet\cite{deng2009imagenet} pre-trained model on other datasets to determine the transfer performance. Following CoCoOp\cite{zhou2022conditional}, our model is trained on all 1000 ImageNet classes in a few-shot manner.\\
\textbf{Domain Generalisation}: We evaluate PromptSync on out-of-distribution (OOD) datasets for domain generalizability. Similar to cross-dataset, we evaluate our ImageNet-trained model directly on OOD datasets, which are described in Section \ref{sec:exp}.

\section{Performance and Latency}
\label{ablate:platency}
The experiments presented in the table \ref{tab:platency} above involve a comparison of different methods, namely MaPLe + TPT, PromptAlign, PromptSync*, and PromptSync. In these experiments, we evaluated the top-1 average accuracy (\%) and latency (in hours for a single prompt update) of each method. Specifically, we investigated PromptSync with and without saving the updated prompt obtained after prototype discrimination, with the variant denoted as PromptSync* indicating the adaptation of prompt tokens for test samples after restoring saved prompt tokens.

The results, as shown in Table \ref{tab:platency}, include latency measurements represented in hours for a single prompt update, and all evaluations are conducted on the ImageNet-A dataset. Notably, the PromptSync* variant demonstrates a faster processing time compared to the full PromptSync method, with only a marginal drop in performance. This outcome underscores the achieved generalization through prototype alignment. Furthermore, in comparison to previous methods such as MaPLe + TPT and PromptAlign, the PromptSync* variant exhibits only a slight increase in latency (0.03 hours) while still improving overall performance.
\section{Sensitivity Comparison}
\label{sensitivity}
We further performed the sensitivity comparison of our method as compared to other state-of-the-art baselines. Figure \ref{fig:tradeoff}(a) shows the comparison of performance during test time adaptation as the number of views increases. All the
results are on ImageNet-A dataset. In comparison to PromptAlign and MaPLe + TPT, their performance almost plateaus around 64 views with insignificant improvement further, while PromptSync shows a consistent improvement with the increase in views and insignificant improvement beyond 128. This proves the generalizability achieved by our method since it optimises base CLIP over a larger number of possible shifts in the dataset, resulting in better performance. Figure \ref{fig:tradeoff}(b) shows the performance comparison as the number of prompt update steps increases. All the methods increase their performance with an increase in the number of steps; however, our method shows better adaptation to the test sample with more steps in comparison to PromptAlign and MaPLe + TPT. For apples-to-apples comparison we perform a single-step update (with 128 views) following TPT \cite{shu2022test}. 

\section{LAION400M Proxy Dataset Analysis}
\label{laion}
Given CLIP's impressive zero-shot performance on ImageNet, we opted for ImageNet as a viable proxy source dataset, aligning with prior research \cite{samadh2023align}. We worked with a subset of LAION400M, comprising 2.5 million images (2 times the size of ImageNet). Furthermore, we carried out an ablation study on the alignment strategy using LAION400M as the source dataset, a dataset known to mirror CLIP's training dataset \cite{cherti2023reproducible}. The results for this ablation study is shown in Table \ref{tab:laion}. Notably, the performance impact remains consistent when utilizing this subset of LAION400M alongside ImageNet. Source class prototypes are computed on the proxy source data to derive the distribution for alignment during test time. As this proxy dataset aligns with the model's training set, this offline computation remains unchanged despite environmental shifts and only necessitates computation once.

\end{document}